\documentclass[conference]{IEEEtran}
\IEEEoverridecommandlockouts
\usepackage{url}
\usepackage{cite}
\usepackage{amsmath,amssymb,amsfonts}
\usepackage{algorithmic}
\usepackage{algorithm}
\usepackage{graphicx}
\usepackage{textcomp}
\usepackage{xcolor}
\usepackage{booktabs}
\usepackage{subfigure}
\def\BibTeX{{\rm B\kern-.05em{\sc i\kern-.025em b}\kern-.08em
    T\kern-.1667em\lower.7ex\hbox{E}\kern-.125emX}}
\begin{document}

\title{Collaborative Label Correction via \\Entropy Thresholding}

\author{\IEEEauthorblockN{
Hao Wu, Jiangchao Yao, Jiajie Wang, Yinru Chen, Ya Zhang, Yanfeng Wang}
\IEEEauthorblockA{Cooperative Medianet Innovation Center \\
Shanghai Jiao Tong University\\
\{howiethepeanut, sunarker, ww1024, chenyinru990504, ya\_zhang, wangyanfeng\}@sjtu.edu.cn}}


\maketitle

\begin{abstract}
  Deep neural networks (DNNs) have the capacity to fit extremely noisy labels nonetheless they tend to learn data with clean labels first and then memorize those with noisy labels. We examine this behavior in light of the Shannon entropy of the predictions and demonstrate the low entropy predictions determined by a given threshold are much more reliable as the supervision than the original noisy labels. It also shows the advantage in maintaining more training samples than previous methods. Then, we power this entropy criterion with the Collaborative Label Correction (CLC) framework to further avoid undesired local minimums of the single network. A range of experiments have been conducted on multiple benchmarks with both synthetic and real-world settings. Extensive results indicate that our CLC outperforms several state-of-the-art methods.
\end{abstract}


\section{Introduction}
Large-scale supervised training datasets have significantly driven the success of DNNs in multiple fields~\cite{krizhevsky2012imagenet,he2016deep,sutskever2014sequence}.
However, accurate labels provided by the human experts are not always feasible in the real-world applications.
Alternative ways of obtaining annotations~\cite{raykar2010learning,fergus2010learning} inevitably introduce label noise to the dataset. As DNNs are capable of fitting extremely noisy labels~\cite{zhang2016understanding}, training DNNs robustly with noisy supervision~\cite{yao2019safeguarded,yao2018deep,han2018masking,cheng2017learning} has been an important and vibrant area of research.
Recent works have proposed two following popular ways for learning with noisy labels. One way is to find the clean instances out of the noisy ones for training via \emph{sample selection}. For example, Decouple~\cite{malach2017decoupling} selects samples on which the two models make different predictions. Co-teaching~\cite{han2018co} uses two models to select small loss samples as supervision for each other. The other way is by leveraging the predictions of the model to rectify the noisy labels, i.e., \emph{label correction}. For instance, Distillation~\cite{li2017learning} combines the predictions of the teacher net with the original labels as supervision for the student net. Co-distillation~\cite{anil2018large,song2018collaborative} collaboratively distills knowledge between the two networks. 

However, for current label correction methods, the predictions of the model are rigidly utilized without verifying their validity. The performance still suffers from the influence of the inaccurate supervision. 
\begin{figure} 
\centering
\includegraphics[width=0.45\textwidth]{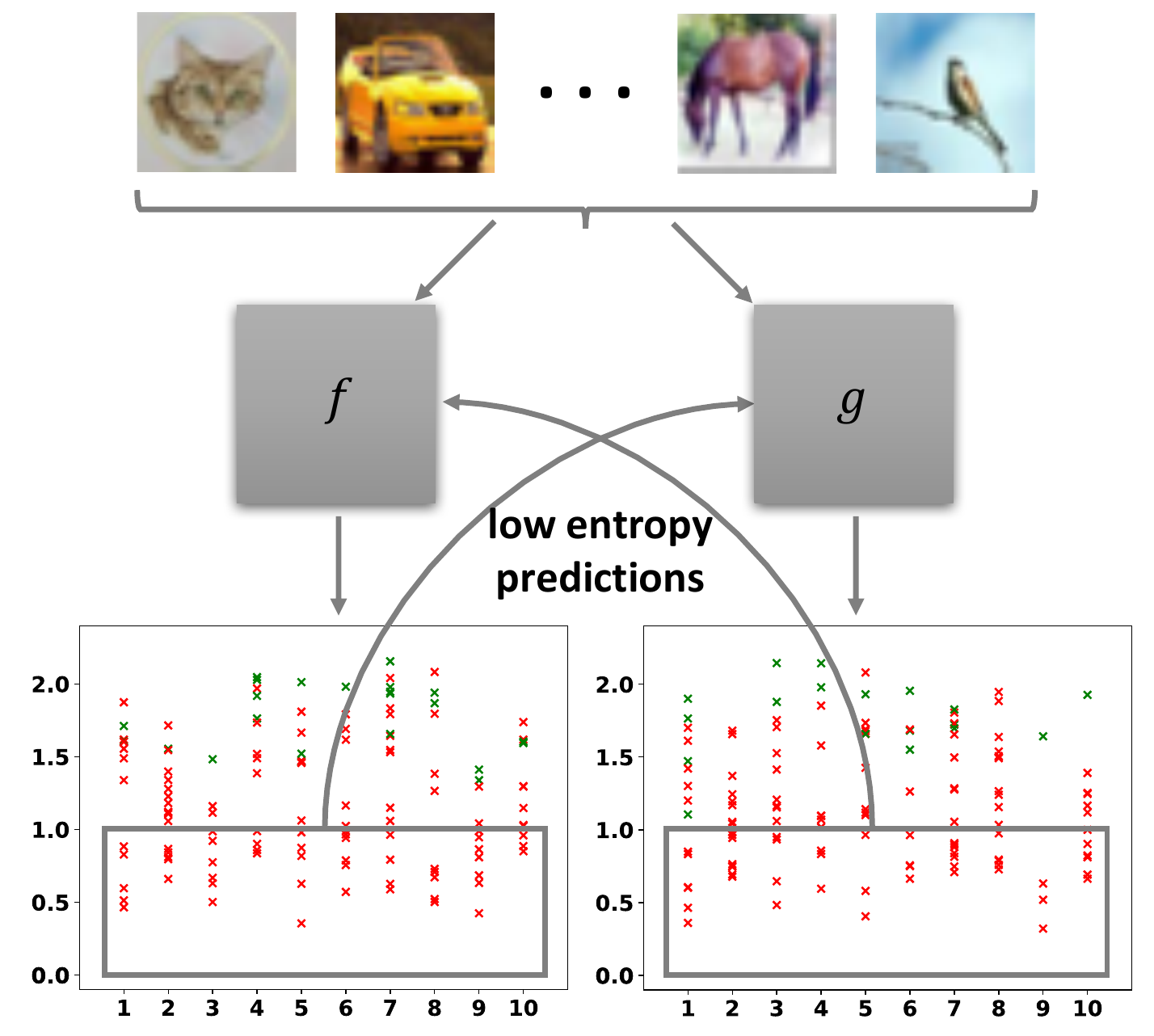}
\caption{The framework of Collaborative Label Correction. We maintain two networks $f$ and $g$. On the panels below, the x-axis represents ten classes in CIFAR-10 and the y-axis represents the entropy of the predictions. We plot the entropy of predictions in a mini-batch as red crosses which indicate correct predictions and green crosses which indicate incorrect predictions.}\label{fig:clc_framework}
\end{figure}
On the other hand, although sample selection methods use different criteria to guarantee the label precision of the remaining samples, the increase of noise ratio will result in the decreasing number of samples. Thus sample selection methods are naturally forced to discard training samples that could be potentially utilized to boost the performance of the classifier. This presents a challenge to learning with noisy labels---can we leverage more reliable supervisions from the noisy dataset while maintain sufficient samples for training?   

To tackle the issues of both ways, we first provide a new method that selects trustworthy predictions to correct the noisy labels, and then propose the Collaborative Label Correction (CLC) framework shown in Fig.~\ref{fig:clc_framework}. As can be seen, two models utilize their low entropy predictions determined by a given threshold to correct labels for each other. We empirically demonstrate the entropy criterion that stems from the memorization effects~\cite{arpit2017closer}, can guarantee the validity of the predictions and thus avoid the poor quality predictions. Treating the low entropy predictions as pseudo labels, CLC manages to achieve better or comparable label precision with sample selection methods. Moreover, CLC shows advantages in keeping much more training samples compared to sample selection methods which averts the loss of data. Simultaneously, CLC greatly benefits from the dual network structure that avoids undesired local minimums of the single network. Our whole model is trained in an end-to-end manner. 

The main contributions are summarized as follows:
\begin{itemize}
    \item We propose entropy thresholding to select the low entropy predictions as the supervision of the classifier. It guarantees the validity of low entropy predictions leveraging the memorization effects~\cite{arpit2017closer} and simultaneously maintains sufficient training samples.
    \item We propose a Collaborative Label Correction framework that incorporates the above advantages of the entropy criterion. Moreover, the dual network structure of CLC avoids undesired local minimums of the single network.
    \item 
    We empirically verify the proposed method on on multiple benchmarks with both synthetic and real-world settings. Comprehensive experiments demonstrate that CLC outperforms several state-of-the-art methods.
     
\end{itemize}

\section{Collaborative Label Correction}
For noisy supervision, consider a set $\mathcal{D}=\{(x_n, y_n)\}^N_{n=1}$ of $N$ samples from $c$ classes, where $x_n$ denotes an image and $y_n\in \{0,1\}^c$ is the corresponding noisy label. To address the issue of inaccurate predictions in label correction methods and loss of data in sample selection methods, our goal is to select the most trustworthy predictions as pseudo labels and retain as many reliable predictions as possible in the training.

\subsection{Memorization Effects in Light of Shannon Entropy}
\begin{figure} 
\centering
\includegraphics[width=0.45\textwidth]{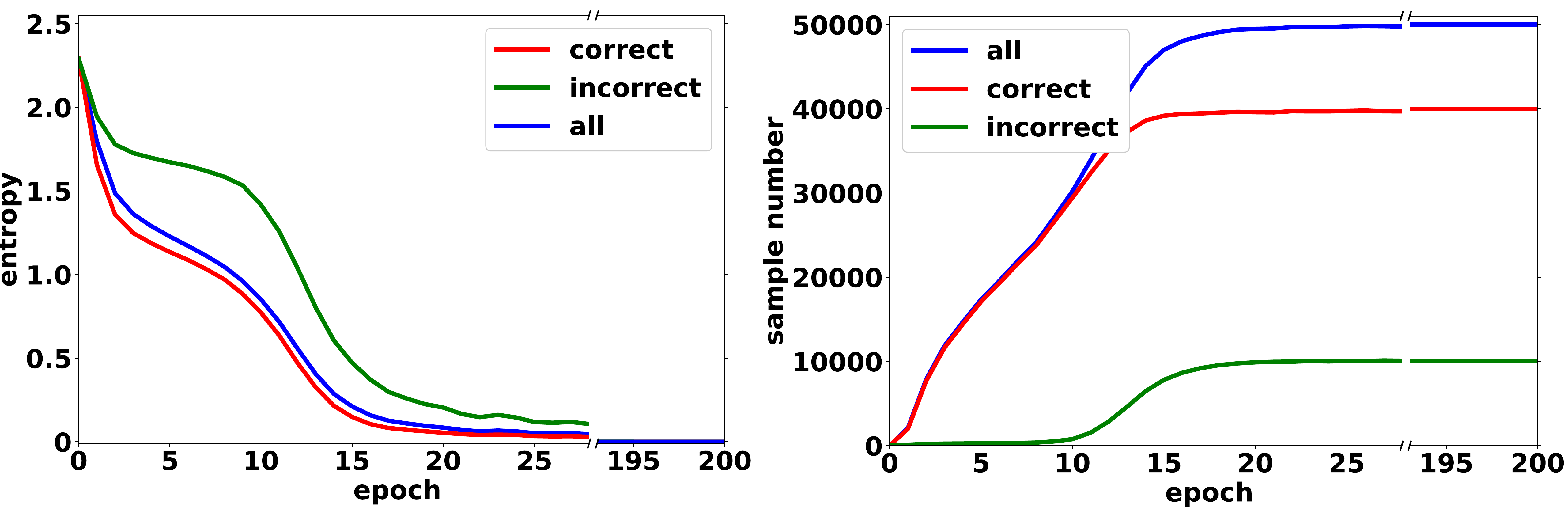}
\caption{\textbf{Left: }The evolution of the average entropy of correct predictions, incorrect predictions and all predictions through the training process. \textbf{Right: }Number of samples with the low entropy predictions determined by a given threshold vs. number of epochs.  }\label{fig:evolution}
\end{figure}

In this section, we will demonstrate that entropy thresholding can help us find reliable predictions out of all model predictions. In a standard training process, DNNs tend to learn samples with clean labels first and then memorize those with noisy labels, which is regarded as the memorization effects~\cite{arpit2017closer}. We examine this process in light of Shannon entropy. For a given input $x$ and a network $f$, $p_f=f(x)$ is denoted as a probabilistic prediction after the softmax output. Its Shannon entropy is given by $H(f,x) = - \sum_{i=1}^c p_{fi} \log p_{fi}$.
Generally, at the beginning of training the model does not learn anything thus the prediction distribution is uniformly distributed which yields the highest entropy. While in an advanced stage of training, the model fits all the labels thus the predictions place all probability in one class for every sample which yields the lowest entropy. To visualize the prediction entropy in the training process, we conduct the toy experiments on CIFAR-10 with the synthetic symmetric noise~\cite{van2015learning}
and depict the corresponding curve on the left panel of Fig.~\ref{fig:evolution}. Besides, we plot the variation of the entropy for both correct predictions and incorrect predictions, since we have the ground-truth labels. As can be seen in the training process, the entropy of predictions generally decreases to zero. The entropy of the correct predictions decreases fast while the entropy of the incorrect predictions decreases at a rather slow pace. The entropy gap between the correct and the incorrect predictions becomes larger at first then narrows down which matches the behavior of the DNNs stated in the memorization effects~\cite{arpit2017closer}. 

If we only analyze the low entropy predictions under a given threshold $1.0$, we will obtain the curve on the right panel of Fig.~\ref{fig:evolution}. We can see that the number of the correct predictions increases sharply and persists at a high level. Conversely, the number of the incorrect predictions remains at a low level for a period and then increases as the training proceeds. This demonstrates the low entropy predictions are trustworthy before the model gradually fits noise. Utilizing this behavior, we propose entropy thresholding to discriminate the reliable low entropy predictions from the unreliable high entropy predictions. We further leverage the low entropy predictions to robustly train our classifier.

\subsection{The Collaborative Label Correction Framework}

In this section, we propose the Collaborative Label Correction (CLC) framework where two models collaboratively utilize their low entropy predictions to correct the original labels for each other. For a given threshold $\Gamma$, we firstly obtain the low entropy set as well as the high entropy set from a given mini-batch $\overline{\mathcal{D}}$ out of $\mathcal{D}$:
\begin{align}
    &\overline{\mathcal{D}}_{f,low} = \{(x_n,p_{fn})|H(f,\overline{\mathcal{D}}) \leq \Gamma\},\label{eq:Dflow}\\
    &\overline{\mathcal{D}}_{g,low} = \{(x_n,p_{gn})|H(g,\overline{\mathcal{D}})\leq \Gamma\},\label{eq:Dglow}\\
    &\overline{\mathcal{D}}_{f,high} = \{(x_n,y_n)|H(f,\overline{\mathcal{D}}) > \Gamma\},\label{eq:Dfhigh}\\
    &\overline{\mathcal{D}}_{g,high} = \{(x_n,y_n)|H(g,\overline{\mathcal{D}})> \Gamma\},\label{eq:Dghigh}
\end{align}

with respect to the predictions from the two networks $f$ and $g$. Note that we only correct labels with model predictions in the low entropy set and leave the labels unchanged in the high entropy set. The objective functions for the two networks $f$ and $g$ are defined as follows: 

\begin{align}
    L(f,\overline{\mathcal{D}}) =& L_{ce}(f,\overline{\mathcal{D}}_{g,low}) + \alpha H(f,\overline{\mathcal{D}}_{f,low}) + \nonumber\\
    + & \beta  L_{ce}(f,\overline{\mathcal{D}}_{g,high}),\label{eq:Lf}\\
    L(g,\overline{\mathcal{D}}) =& L_{ce}(g,\overline{\mathcal{D}}_{f,low}) + \alpha H(g,\overline{\mathcal{D}}_{g,low}) + \nonumber\\
    + & \beta  L_{ce}(g,\overline{\mathcal{D}}_{f,high}),\label{eq:Lg}
\end{align}

\begin{algorithm}[tb]
 \scriptsize
\caption{The CLC Algorithm}
\label{alg:clcalgorithm}
\begin{algorithmic}[1] 
\REQUIRE A noisy dataset $\mathcal{D}$, two networks $f$ and $g$, $\Gamma$, $\alpha$, $\beta$ 
\FOR{epoch $i=\text{StartEpoch}$ to $\text{MaxEpoch}$ }
\FOR{batch $j=1$ to $\frac{|\mathcal{D}|}{\text{BatchSize}}$ }
\STATE Calculate $H(f,\overline{\mathcal{D}})$
\STATE Calculate $H(g,\overline{\mathcal{D}})$
\STATE Obtain $\overline{\mathcal{D}}_{f,low}$ and $\overline{\mathcal{D}}_{f,high}$ by Eq.~\eqref{eq:Dflow} and Eq.~\eqref{eq:Dfhigh}
\STATE Obtain $\overline{\mathcal{D}}_{g,low}$ and $\overline{\mathcal{D}}_{g,high}$ by Eq.~\eqref{eq:Dglow} and Eq.~\eqref{eq:Dghigh}
\STATE Update $f$ by optimizing $L(f,\overline{\mathcal{D}})$ in Eq.~\eqref{eq:Lf}
\STATE Update $g$ by optimizing $L(g,\overline{\mathcal{D}})$ in Eq.~\eqref{eq:Lg}
\ENDFOR
\ENDFOR
\end{algorithmic}
\end{algorithm}

where $L_{ce}$ is the cross-entropy loss. In the first term of Eq.~\ref{eq:Lf}, the network $f$ is supervised by the low entropy predictions from network $g$, which is the key module of this paper. The second term of Eq.~\ref{eq:Lf} is a regularization term to prevent the output of the network $f$ from degenerating to the uniform distribution. We minimize this term in order to make $f$ maintain the already reliable predictions. For the third part of Eq.~\ref{eq:Lf}, it is an auxiliary term that guarantees $f$ to use the original labels in the early phase, since the number of low entropy predictions in this stage is small. As the training proceeds, the number of low entropy predictions will grow and conversely the third term will be squeezed out. Like~\cite{song2018collaborative}, we empirically assign weights to $\alpha$ and $\beta$ to save human labor. The training process is summarized in Algorithm~\ref{alg:clcalgorithm}.

\subsection{Prediction Selection vs. Sample Selection}
In this section, we present the advantage of prediction selection in CLC over sample selection. When treating all model predictions as pseudo labels, prediction selection which selects samples with the correct pseudo labels seems similar to sample selection which selects samples with correct original labels. 
However, for sample selection, the optimal situation is that all of the clean instances in the dataset are selected and the noisy instances are discarded, thus the upper bound of the number of selected samples is equal to the amount of clean samples collected in the original dataset. Conversely, for prediction selection, we can easily obtain a suboptimal model that simply fits all the labels according to the memorization effects~\cite{arpit2017closer}. Thus the upper bound of the number of correct pseudo labels is no less than the amount of the correct original labels. The optimal situation is that a well generalized model corrects all the labels and the least upper bound is the total amount of samples. Namely, prediction selection has the potential to maintain all the training samples including the originally noisy ones. Conversely, sample selection is forced to discard training samples that could be potentially leveraged to enhance the performance of the classifier. 

\subsection{Dual Network vs. Single Network}
We choose to maintain two networks collaboratively for CLC. Naturally a question would occur that if we can use only one network to learn from its own low entropy predictions, namely Self-paced Label Correction (SLC). However, there reside two problems as follows. 
One problem is that although entropy thresholding can guarantee the validity of most predictions, the few errors from the prediction bias can not be filtered by their own predictor. Thus the errors will accumulate through training. This will cause the self-paced network to be stuck in \emph{undesired local minimums} as soon as all of the labels are corrected by the low entropy predictions. CLC naturally addresses this problem as two networks have the potential to filter out the biases residing in each other.
The other problem is that stochastic tweaking may cause some of the correct low entropy predictions to go wrong even though we explicitly encourage the model to minimize the entropy of the already low entropy predictions. These alterations are presumably irreversible. Take ``gene mutation" as a supportive example that we have only one good genotype corresponding to the correct category and much more bad genotypes corresponding to the incorrect categories. The rate of mutation will be much higher than that of reverse mutation. Since one model is not aware of the prediction errors of its own, the accumulation of the ``mutations" may cause SLC to collapse into \emph{undesired local minimums} where all predictions are random low entropy predictions. For CLC, incorrect predictions caused by non-trivial tweaking will not be directly used as supervision for the original predictor. Thus CLC avoids the accumulation of the ``mutations" and enhances the stability of label correction. 

\section{Experiments}
\begin{table}
\tiny
\centering
\caption{Summary of the 6 datasets used in the experiments.}
\label{tab:datasets}
\begin{tabular}{lccccc}  
\toprule
\textbf{dataset} & \# of training & \# of testing & \# of class & data type & noise type \\
\midrule
CIFAR-10 & 50,000 & 10,000 & 10 & image & synthetic\\
CIFAR-100 & 50,000 & 10,000 & 100 & image & synthetic\\
Clothing1M & 1M & 74k & 14 & image & real-world\\
Food-101N & 310k & 25,250 & 101 & image & real-world\\
20 Newsgroups & 11,314 & 7,532 & 20 & text & synthetic\\
TREC & 5,452 & 500 & 6 & text & synthetic\\
\bottomrule
\end{tabular}

\end{table}

\subsection{Datasets and Baselines}\label{sec:datasets}
\begin{table*}
\tiny
\centering
\caption{Average test accuracy (\%) on CIFAR-10 \& CIFAR-100 over the last ten epochs.}
\label{tab:cifar}
\begin{tabular}{lrrrrrr|rrrrrr}  
\toprule
 & \multicolumn{6}{c}{\textbf{CIFAR-10}} & \multicolumn{6}{c}{\textbf{CIFAR-100}}\\
\midrule
 & \multicolumn{2}{c}{\textbf{pairwise}} & \multicolumn{2}{c}{\textbf{asymmetric}} & \multicolumn{2}{c}{\textbf{symmetric}} & \multicolumn{2}{c}{\textbf{pairwise}} & \multicolumn{2}{c}{\textbf{asymmetric}} & \multicolumn{2}{c}{\textbf{symmetric}}\\
 \midrule
\textbf{setting} & 1 & 2 & 3 & 4 & 5 & 6 & 7 & 8 & 9 & 10 & 11 & 12\\
\midrule
\textbf{noise ratio} & 0.2 & 0.45 & 0.2 & 0.45 & 0.2 & 0.5 & 0.2 & 0.45 & 0.2 & 0.45 & 0.2 & 0.5\\
\midrule
Standard & 76.54 & 49.73 & 82.86 & 69.52 & 76.71 & 49.91 & 50.47 & 32.51 & 50.79 & 31.47 & 48.63 & 25.44\\
Bootstrap & 76.19 & 49.75 & 82.90 & 70.23 & 76.66 & 48.83 & 51.01 & 31.17 & 50.35 & 32.42 & 47.87 & 24.74\\
Forward & 78.06 & 58.69 & 83.29 & 69.59 & 77.56 & 51.70 & 52.67 & 30.14 & 54.99 & 28.41 & 47.00 & 25.94\\
Decouple & 77.30 & 49.23 & 83.53 & 70.24 & 78.03 & 48.97 & 51.28 & 31.49 & 50.50 & 31.72 & 46.19 & 23.29\\
MentorNet & 80.32 & 58.67 & 83.62 & 70.28 & 80.73 & 68.70 & 50.89 & 32.26 & 50.27 & 32.35 & 52.12 & 37.96\\
Co-teaching & 83.24 & 72.74 & 85.12 & 76.02 & 82.09 & 74.06 & 54.74 & 34.08 & 52.81 & 34.77 & 53.84 &  \textbf{41.34}\\
Co-distillation & 81.36 & 51.87 & 85.28 & 72.15 & 81.49 & 56.76 & 56.40 & 34.93 & 55.27 & 35.27 & 54.36 & 32.41\\
CLC & \textbf{88.97} & \textbf{84.26} & \textbf{89.71} & \textbf{78.29} & \textbf{85.73} & \textbf{80.97} & \textbf{62.12} & \textbf{54.62} & \textbf{62.37} & \textbf{55.46} & \textbf{57.45} & 41.17\\
\bottomrule
\end{tabular}
\end{table*}

The general information of the datasets can be found in Table~\ref{tab:datasets}. 
For CIFAR10 and CIFAR100~\cite{krizhevsky2009learning}, 
pairwise noise~\cite{han2018co}, asymmetric noise~\cite{patrini2017making} and symmetric noise~\cite{van2015learning} are injected. 
We use $r$ to denote the noise ratio.
Images in Clothing1M~\cite{xiao2015learning} are crawled from online shopping websites with labels generated by surrounding texts. The overall label accuracy is 61.54\%.
Food-101N~\cite{lee2018cleannet} is generated through Google, Bing, Yelp and TripAdvisor.
The overall label accuracy is estimated to be 80\%. 
20 Newsgroups~\cite{20newsgroup} is a collection of 20 newsgroups and TREC~\cite{li2002learning} is a collection of questions from 6 coarse classes. 
Pairwise noise and symmetric noise are injected.
We compare CLC with the following methods: {\bf Standard}, which directly trains a vanilla classifier on noisy datasets; {\bf Bootstrap}~\cite{reed2014training}, which linearly combines the model predictions and original labels; {\bf Forward}~\cite{patrini2017making}, which uses a noise transition matrix to modify model predictions; {\bf Decouple}~\cite{malach2017decoupling}, which updates the parameters when two models disagree with each other; 
{\bf MentorNet}~\cite{jiang2018mentornet}. We use self-paced MentorNet that the model learns with its small loss samples; {\bf Co-teaching}~\cite{han2018co}, which uses two models to select small loss samples for each other.; {\bf Co-distillation}~\cite{anil2018large,song2018collaborative}, which uses two models to distill knowledge for each other.



\subsection{Implementation} \label{sec:implemt}
For CIFAR-10 and CIFAR-100, we align the implementation in~\cite{han2018co} that a 9-layer CNN is used as the classifier. 
We generally deploy the Adam optimizer~\cite{kingma2015adam} and set the learning rate to 0.001. We set the batch size to 128 and run 200 epochs in total.
For 20 Newsgroups and TREC, a 3-layer MLP is used and the dimension of the hidden layers is set to 300. We use the pre-trained word embeddings from GloVe~\cite{pennington2014glove}. 
For Clothing1M and Food-101N, ResNet-50~\cite{he2016deep} is used. 
Images are resized to 256 with respect to shorter sides. We perform random crop of 224$\times$224, random flip, brightness, contrast and saturation. We set the batch size to 64 and run 15 epochs on Clothing1M and 30 epochs on Food-101N.
To be fair, all methods are pre-trained for several epochs as warming-up following~\cite{patrini2017making, song2018collaborative, han2018co}. 
Specifically the models are trained as Standard for 2 epochs on Clothing1M and 10 epochs on all other datasets. 
For Forward, the pre-trained model is used to estimate the transition matrix~\cite{patrini2017making} on all datasets except for Clothing1M where we use the normalized ground-truth confusion matrix provided in~\cite{xiao2015learning}. For MentorNet and Co-teaching, the noise ratio $r$ is required as side information~\cite{han2018co}.
For CLC, we empirically use the average entropy of the pre-trained model predictions as $\Gamma$. We empirically set $\alpha=0.1$ and $\beta=0.5$. We use test accuracy to measure the performances. For sample selection methods along with CLC, we also report the number of selected samples and their label precision
on the CIFAR datasets as we have the ground-truth labels.

\begin{figure} 
\centering
\includegraphics[width=0.45\textwidth]{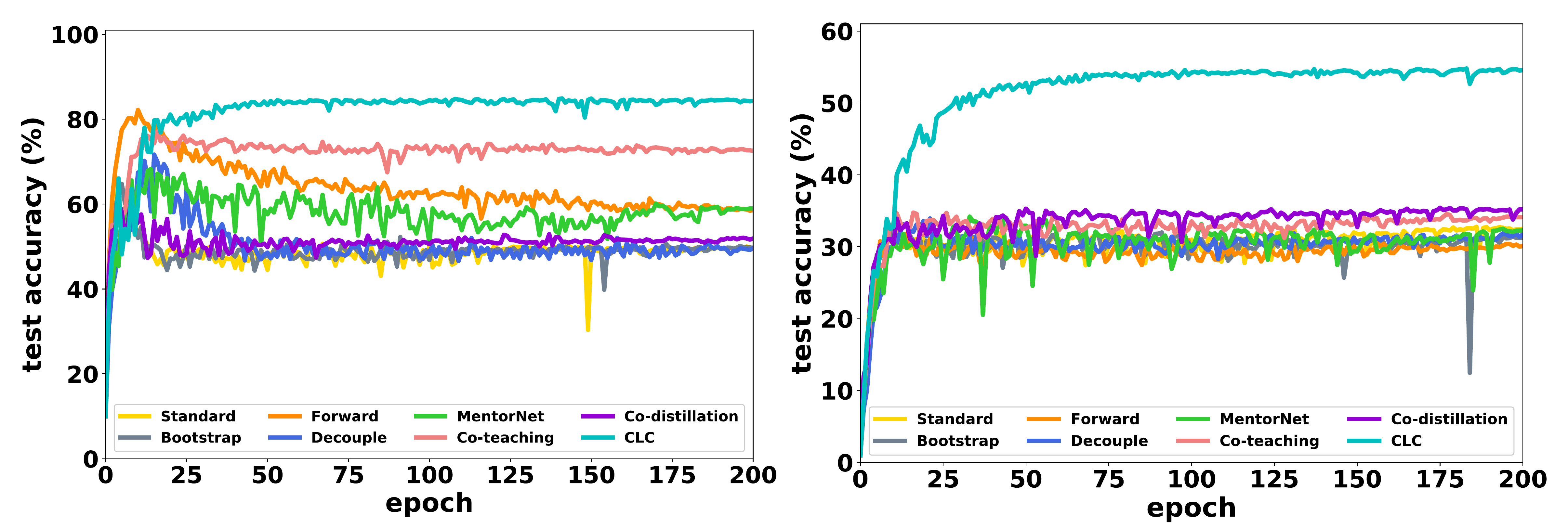}
\caption{\textbf{Left: }Illustration of test accuracy vs. number of epochs on CIFAR-10 with pairwise noise. ($r=0.45$). \textbf{Right: }Illustration of test accuracy vs. number of epochs on CIFAR-100 with pairwise noise. ($r=0.45$).}\label{fig:pn_045}
\end{figure}

\subsection{Results on CIFAR-10 and CIFAR-100}\label{sec:cifar}


Table~\ref{tab:cifar} summarizes the results of all methods on the CIFAR datasets. 
Our CLC achieves the best performance in most of the settings. 
Specifically for high-level pairwise noise ($r=0.45$), CLC outperforms the best baseline by 11.52\% on CIFAR-10 and 19.69\% on CIFAR-100 even impressively surpassing all baselines under low-level noise. For asymmetric noise, CLC manages to outperform the best baseline by 20.19\% when $r=0.45$ on CIFAR-100 and again surpass all baselines under low-level noise. 
In Fig.~\ref{fig:pn_045}, we trace the test accuracy of all baselines and CLC under high-level pairwise noise on CIFAR-10 on the left panel. For all baselines, the test accuracy increases first and then decreases as the training proceeds which matches the memorization effects.
\begin{figure}
\centering
\includegraphics[width=0.45\textwidth]{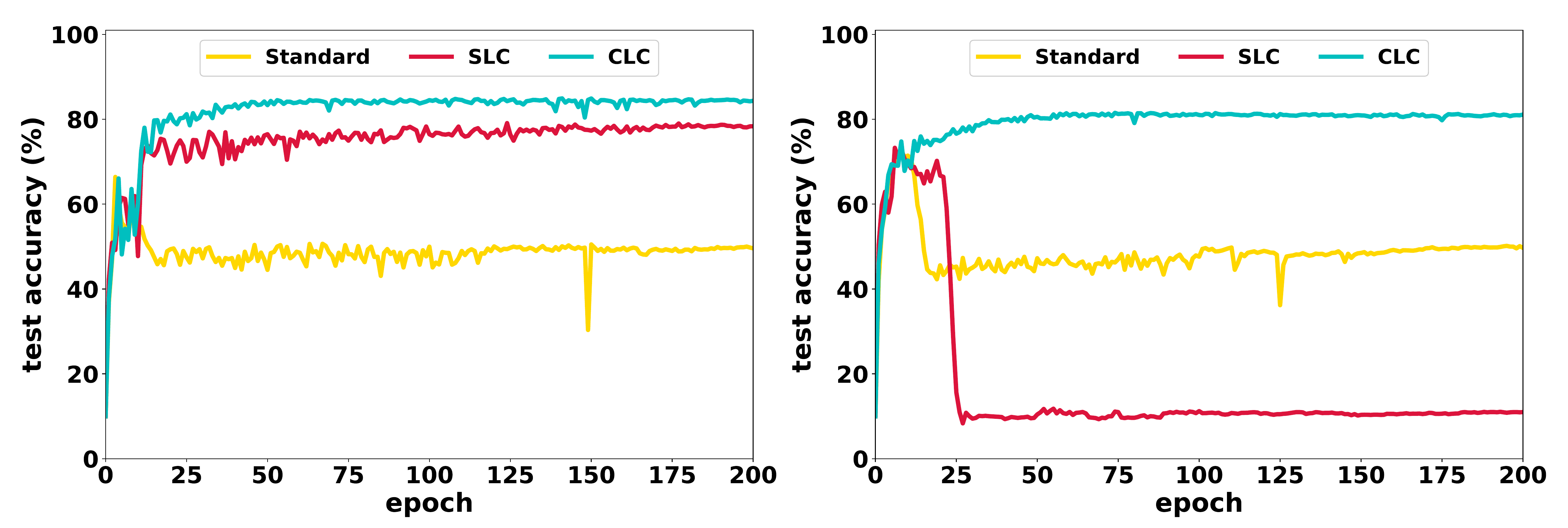}
\caption{\textbf{Left: }Comparison of CLC and SLC on CIFAR-10 with pairwise noise ($r=0.45$). \textbf{Right: }Comparison of CLC and SLC on CIFAR-10 with symmetric noise ($r=0.5$).}\label{fig:slc}
\end{figure}
 However, the accuracy of CLC keeps increasing and persists at a high level after a long time of training, which demonstrates the robustness of CLC under extreme label noise. On the right panel, we trace the test accuracy on CIFAR-100 with high-level pairwise noise. 
 Similarly, CLC keeps improving for around 60 epochs and persists at a high level till the end which indicates CLC is effective in dealing with multiple classes. 

\begin{figure*} 
\centering
\includegraphics[width=0.9\textwidth]{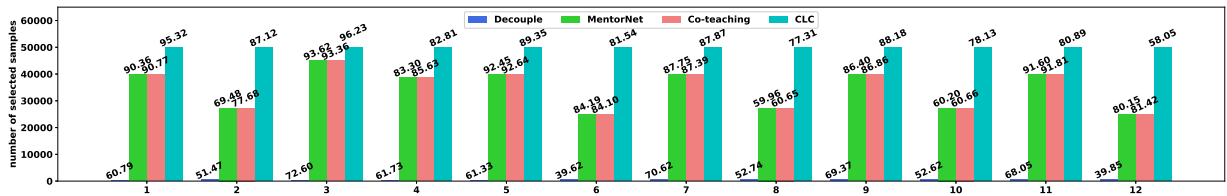}
\caption{Average number of selected samples and average label precision (\%) over the last ten epochs on the CIFAR datasets of 12 different settings. The indices of x-axis corresponds to the settings in Table~\ref{tab:cifar}.}\label{fig:num_prec}
\end{figure*}

In Fig.~\ref{fig:num_prec}, we show the average number of selected samples and average label precision. Decouple only selects a few samples and the label precision is low.
MentorNet and Co-teaching are programmed to discard a proportion $r$ of samples thus almost half of the samples are discarded when $r$ is large. Co-teaching achieves better label precision compared to MentorNet taking advantage of collaborative learning. CLC retains all 50000 of the samples as well as achieving comparable or even better label precision in most of the settings. 
Setting 12 is the only one setting where the label precision of CLC is much lower than that of Co-teaching. However, CLC maintains twice the number of samples than Co-teaching. Thus the relatively low label precision is compensated by the number advantage.

In Fig.~\ref{fig:slc} we compare CLC with SLC. For high-level pairwise noise, the test accuracy of SLC is 78.35\%, which is better than all baselines. Moreover, SLC retains all training samples and shows robustness against label noise which resembles the behavior of CLC. However, CLC addresses the prediction bias in SLC by collaboratively correcting labels for each, therefore CLC outperforms SLC by 5.91\%. 
For high-level symmetric noise, SLC works fine in the early stage and manages to correct 49785 samples out of 50000 samples with label precision of 75.85\% at the 19th epoch. Then, SLC degenerates rapidly within 6 epochs due to the stochastic tweaking and is eventually stuck in an undesired local minimum where all predictions are random low entropy predictions. For CLC, the dual model structure is naturally robust to the stochastic tweaking and can be trained stably.

\begin{figure}

\centering
\includegraphics[width=0.45\textwidth]{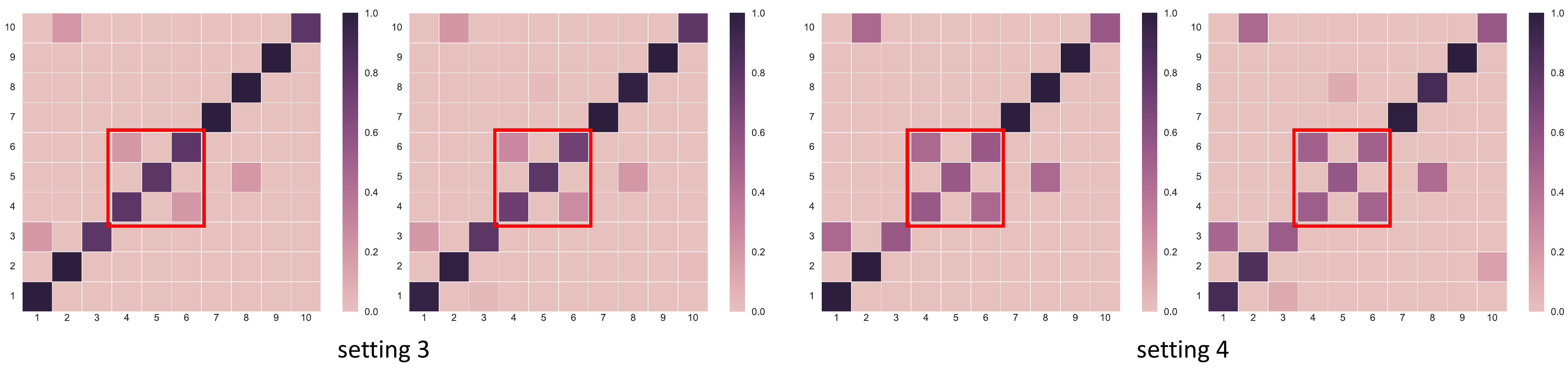}
\label{fig:setting3}

\caption{The ground-truth transition matrices and the confusion matrices of CLC models on CIFAR-10.
The horizontal indices denote the noisy classes while the vertical indices denote the ground-truth classes or predicted classes. 
}
\label{fig:confu_cifar}
\end{figure}

In Fig.~\ref{fig:confu_cifar}, we illustrate the confusion matrices with predictions of CLC models and the noisy labels in setting 3 and 4. The ground-truth transition is given on the left panel in each setting. 
Although CLC does not explicitly model the noise transition, the induced confusion matrices closely resemble the ground-truth transition. Specifically 
labels of the 4th category ``cat'' and 6th category ``dog'' randomly flip between each other forming a hard noise structure visualized as a block (bounded in a red box). Nonetheless the same block can be observed in the confusion matrix. This demonstrates that CLC is capable of capturing hard noise patterns under extreme label noise.

\subsection{Results on Clothing1M and Food-101N}\label{sec:clo_food}

Table~\ref{tab:clothing1m} summarizes the results on two real-world datasets. For Clothing1M, Decouple has a poor performance.
 Co-teaching and MentorNet adopt the same criterion 
 while Co-teaching leverages two models to cope with  the sample selection bias and performs better. Maintaining two models, Co-distillation shows comparable results with Co-teaching. Among all baselines, Forward achieves the best performance with the use of the ground-truth transition matrix. However CLC outperforms Forward by 2.16\% without using any side information. For Food-101N, Bootstrap degenerates the performance with the unverified predictions. Without the knowledge of the ground-truth transition matrix, Forward improves little compared to Standard. 
 Maintaining two networks, Co-distillation shows comparable results with Co-teaching . However, CLC manages to outperform the best baseline by 1.46\%. 
 
\begin{figure} 
\centering
\includegraphics[width=0.45\textwidth]{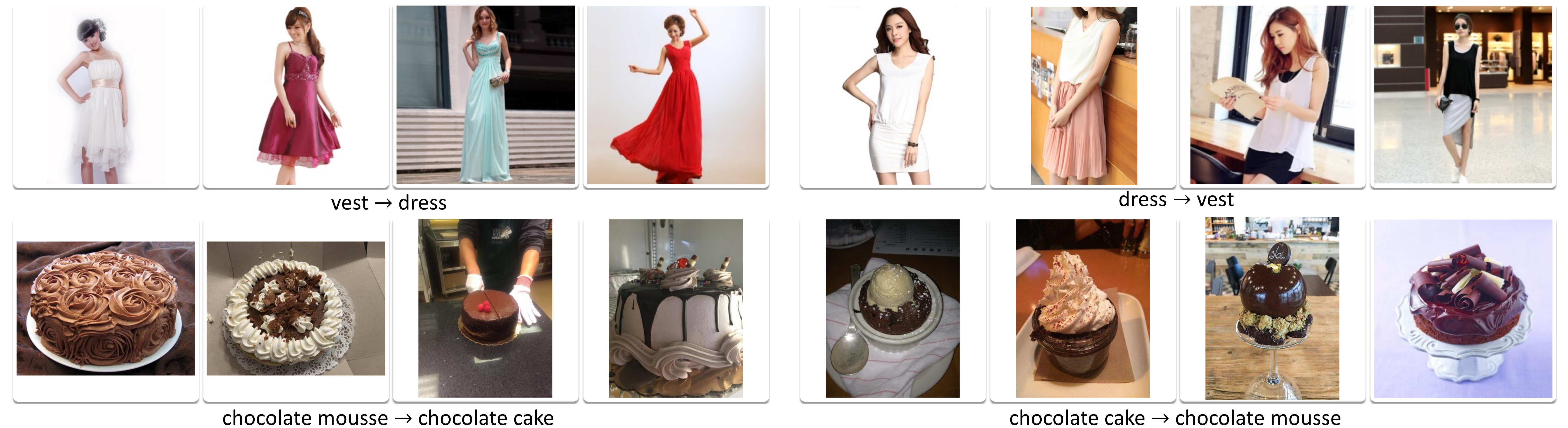}
\caption{Examples of label corrections of CLC on Clothing1M and Food-101N.}\label{fig:examples}
\end{figure}

 In the top panel of Fig.~\ref{fig:examples} we present several examples of label corrections of CLC between two similar classes ``dress'' and ``vest''. Both ``dress'' and ``vest'' cover upper body and lack sleeves while only ``dress'' covers lower body. CLC successfully captures the difference between the two categories and correct the noisy labels with its own predictions.
 In the bottom panel of Fig.~\ref{fig:examples} we present several examples between two similar classes ``cake'' and ``mousse''. They contain the same ingredient but differ in shapes and CLC manages to rectify the incorrect labels. These examples prove that samples with originally noisy labels are potentially useful.


\begin{table}
\centering
\tiny
\caption{Average test accuracy (\%) on Clothing1M and Food-101N over the last five epochs.}
\label{tab:clothing1m}
\begin{tabular}{lrr}  
\toprule
\textbf{method} & Clothing1M & Food-101N \\
\midrule
Standard & 66.14 & 75.45\\
Bootstrap & 66.70 & 74.75\\
Forward & 67.70 & 75.92\\
Decouple & 64.44 & 72.47\\
MentorNet & 60.00 & 76.93\\
Co-teaching & 67.30 & 77.87\\
Co-distillation & 67.23 & 77.31\\
CLC & \textbf{69.86} & \textbf{79.33}\\
\bottomrule
\end{tabular}

\end{table}

\subsection{Results on 20 Newsgroups and TREC}

The results of all methods on 20 Newsgroups and TREC are reported in Table~\ref{tab:news}. Generally, the performance of Standard model is poor especially when noise ratio is high. For label correction methods Bootstrap and Co-distillation, since they heavily rely on the quality of the model predictions, only little improvement is made compared to Standard and failures occur in multiple settings. Forward also leverages the Standard model to estimate the transition matrix thus the performance suffers from the inaccurate estimations in setting 2 and setting 6. Conversely, sample selection methods like Decouple and MentorNet show good robustness against label noise. Adopting the same small loss trick, Co-teaching does not perform as well as MentorNet and failed in setting 2. This indicates the small loss samples selected by one model may not be useful information for the other model. In all settings, CLC achieves the best performance which demonstrates that CLC is also effective in handling textual data under noisy supervision. Impressively, the result of CLC on 20 Newsgroups with high-level symmetric noise even surpasses the results of all baselines under low-level symmetric noise. 
\begin{table}
\tiny
\centering
\caption{Average test accuracy (\%) on 20 Newsgroups over the last ten epochs.}
\label{tab:news}

\begin{tabular}{lrrrr|rrrr}  
\toprule
& \multicolumn{4}{c}{\textbf{20 Newsgroups}} & \multicolumn{4}{c}{\textbf{TREC}}\\
 \midrule
 & \multicolumn{2}{c}{\textbf{pairwise}} &  \multicolumn{2}{c}{\textbf{symmetric}} & \multicolumn{2}{c}{\textbf{pairwise}} &  \multicolumn{2}{c}{\textbf{symmetric}}\\
 \midrule
\textbf{setting} & 1 & 2 & 3 & 4 & 5 & 6 & 7 & 8\\
\midrule
\textbf{noise ratio} & 0.2 & 0.45 & 0.2 & 0.5 & 0.2 & 0.45 & 0.2 & 0.5\\
\midrule
Standard & 53.31 & 40.60 & 50.80 & 30.13 & 70.20 & 44.82 & 70.16 & 50.18\\
Bootstrap & 52.57 & 40.88 & 50.90 & 30.16 & 71.10 & 44.10 & 70.82 & 45.88\\
Forward & 60.55 & 30.69 & 54.94 & 42.26 & 72.22 & 39.20 & 71.86 & 58.22\\
Decouple & 61.15 & 41.34 & 61.28 & 42.12 & 80.02 & 45.62 & 78.12 & 51.22\\
MentorNet & 63.19 & 44.96 & 64.23 & 58.93 & 79.82 & 61.98 & 79.42 & 71.40\\
Co-teaching & 62.82 & 39.98 & 63.48 & 59.83 & 79.78 & 61.82 & 78.14 & 71.54\\
Co-distillation & 55.36 & 38.41 & 51.89 & 32.92 & 72.18 & 47.76 & 70.59 & 49.66\\
CLC & \textbf{66.42} & \textbf{49.24} & \textbf{67.22} & \textbf{65.02} & \textbf{80.81} & \textbf{71.80} & \textbf{80.12} & \textbf{73.42}\\
\bottomrule
\end{tabular}
\end{table}

\begin{figure} 
\centering
\includegraphics[width=0.45\textwidth]{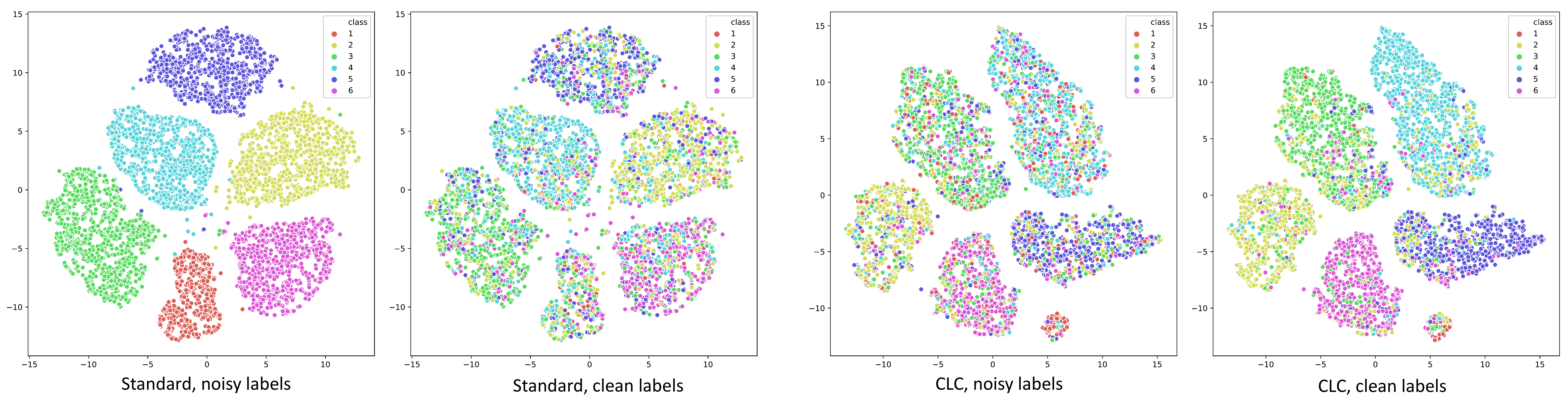}
\caption{The t-SNE visualizations of the last layer before softmax in setting 8. We visualize the feature representation of Standard and CLC with noisy and clean labels superimposed. }\label{fig:tsne}
\end{figure}

In figure~\ref{fig:tsne}, we analyze the learned representation via t-SNE~\cite{maaten2008visualizing}. We visualize the the last layer of the classifier before softmax in setting 8. When the noisy labels are superimposed, the Standard feature representations are well clustered in the view of noisy training labels, which indicates Standard completely fits the noisy supervision. This is consistent with the memorization effects. While for CLC, the clusters contain massive samples from different classes that indicates CLC does not directly rely on the noisy labels. When the true labels are superimposed, clusters of the Standard model turn into mixtures of samples in the view of the ground-truth labels. Conversely, the feature representations of CLC are much better clustered. This demonstrates that CLC can guarantee the model to be trained with reliable supervision under label noise.

\section{Conclusion and Future Work}

In this paper, we present a simple entropy based criterion to select low entropy predictions which are more reliable as supervision than the original noisy labels. Incorporating this entropy based criterion, we propose the CLC framework where two models correct labels for each other with their low entropy predictions. CLC has the potential to maintain much more samples than sample selection methods and the dual network structure of CLC can avoid the undesired local minimums of the single network. 
We conduct a range of experiments on visual and textual datasets with synthetic or real-world noise to demonstrate that CLC trains models robustly even under extreme label noise. 
Future directions of research include automatically adjusting threshold $\Gamma$ by the network itself and other criterion that verifies the validity of predictions for CLC.

\section*{Acknowledgment}
This work is supported by The High Technology Research and Development Program of China (2015AA015801), NSFC (61521062), and STCSM (18DZ2270700).







\bibliographystyle{./bibliography/IEEEtran}
\bibliography{./main_short}

\end{document}